\documentclass[a4paper,conference]{IEEEtran}
\ifCLASSINFOpdf
\else
\fi
\usepackage{amsfonts}

\usepackage{times}
\usepackage{graphicx}
\usepackage{amsmath}
\usepackage{amssymb}
\usepackage{subfigure}

\begin{document}
%
\title{Classic versus deep learning approaches to address computer vision challenges}

\author{\IEEEauthorblockN{Nati Ofir}
\IEEEauthorblockA{Kingston University\\London, UK\\
	Email: natiofir@gmail.com}
\and
\IEEEauthorblockN{Jean-Christophe Nebel}
\IEEEauthorblockA{Kingston University\\London, UK\\Email: j.nebel@kingston.ac.uk}
}


%


\maketitle

\begin{abstract}
Computer vision and image processing address many challenging applications. While the last decade has seen deep neural network architectures revolutionizing those fields, early methods relied on 'classic', i.e., non-learned approaches.	
In this study, we explore the differences between classic and deep learning (DL) algorithms to gain new insight regarding which is more suitable for a given application. The focus is on two challenging ill-posed problems, namely faint edge detection and multispectral image registration, studying recent state-of-the-art DL and classic solutions.
While those DL algorithms outperform classic methods in terms of accuracy and development time, they tend to have higher resource requirements and are unable to perform outside their training space. Moreover, classic algorithms are more transparent, which facilitates their adoption for real-life applications. As both classes of approaches have unique strengths and limitations, the choice of a solution is clearly application dependent.
\end{abstract}


%
\IEEEpeerreviewmaketitle

\section{Introduction} \label{sec:intro}

Computer vision and image processing address many challenging applications. While the last decade has seen deep neural network architectures revolutionizing those fields, early methods relied on 'classic' approaches. Here, 'classic' refers to techniques that do not rely on machine learning, such as engineered feature descriptors, theoretic-based algorithms, search methods, and usage of theoretically proven characteristics. In this study, we explore the differences between classic and deep learning (DL) approaches and their associated constraints in order to gain new insight regarding which is more suitable for a given application. 
While DL is only a subset of machine learning, this manuscript does not cover other machine learning algorithms as they have become less popular. Indeed, currently, around 25\% of all papers presented at computer vision and image processing conferences take advantage of DL. Moreover, a session dedicated to it has become the norm on the program of many scientific venues.

\begin{figure*}[tbh]
\centering
\subfigure[]{\includegraphics[width=0.32\linewidth]{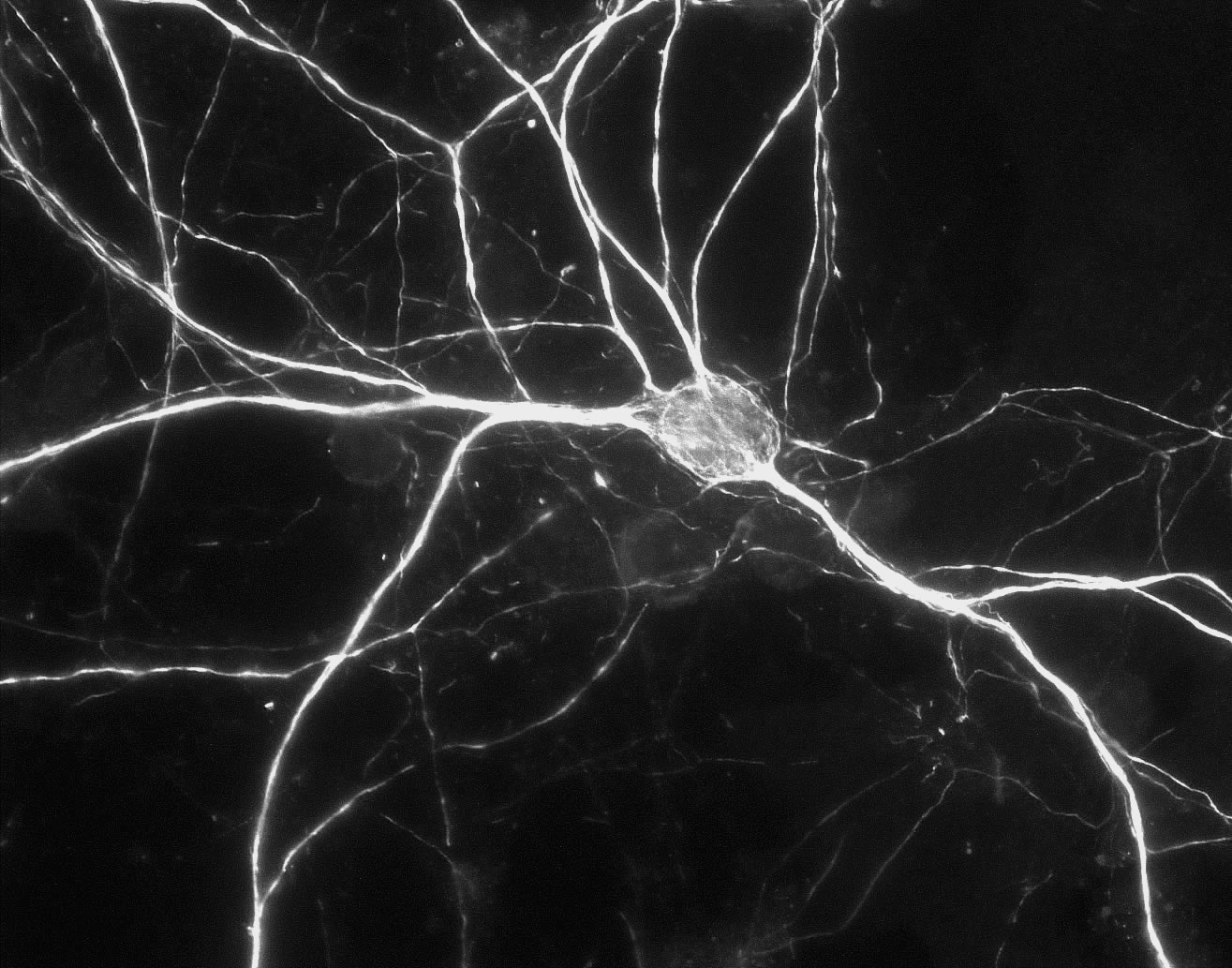}} %
\subfigure[]{\includegraphics[width=0.32\linewidth]{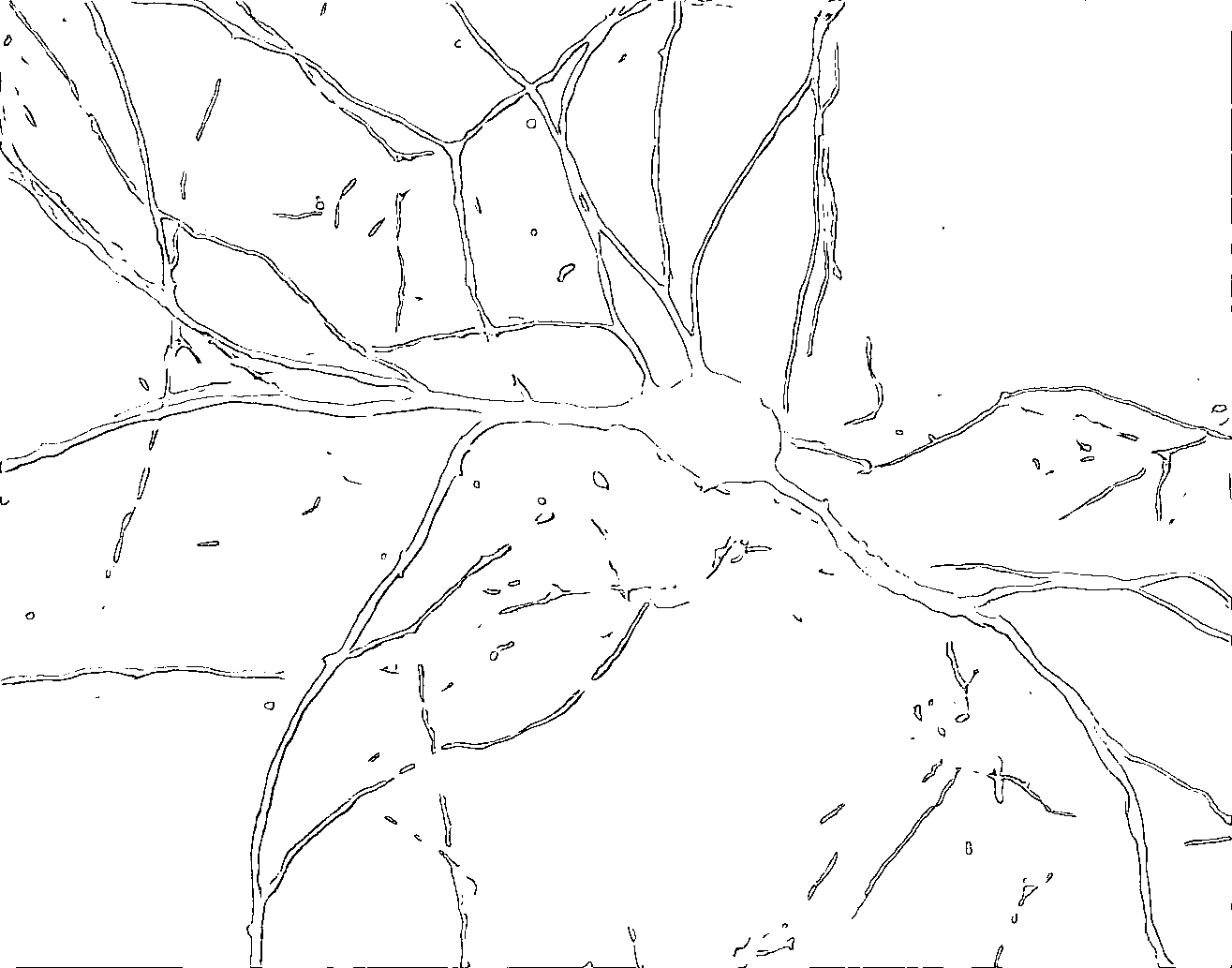}} %
\subfigure[]{\includegraphics[width=0.32\linewidth]{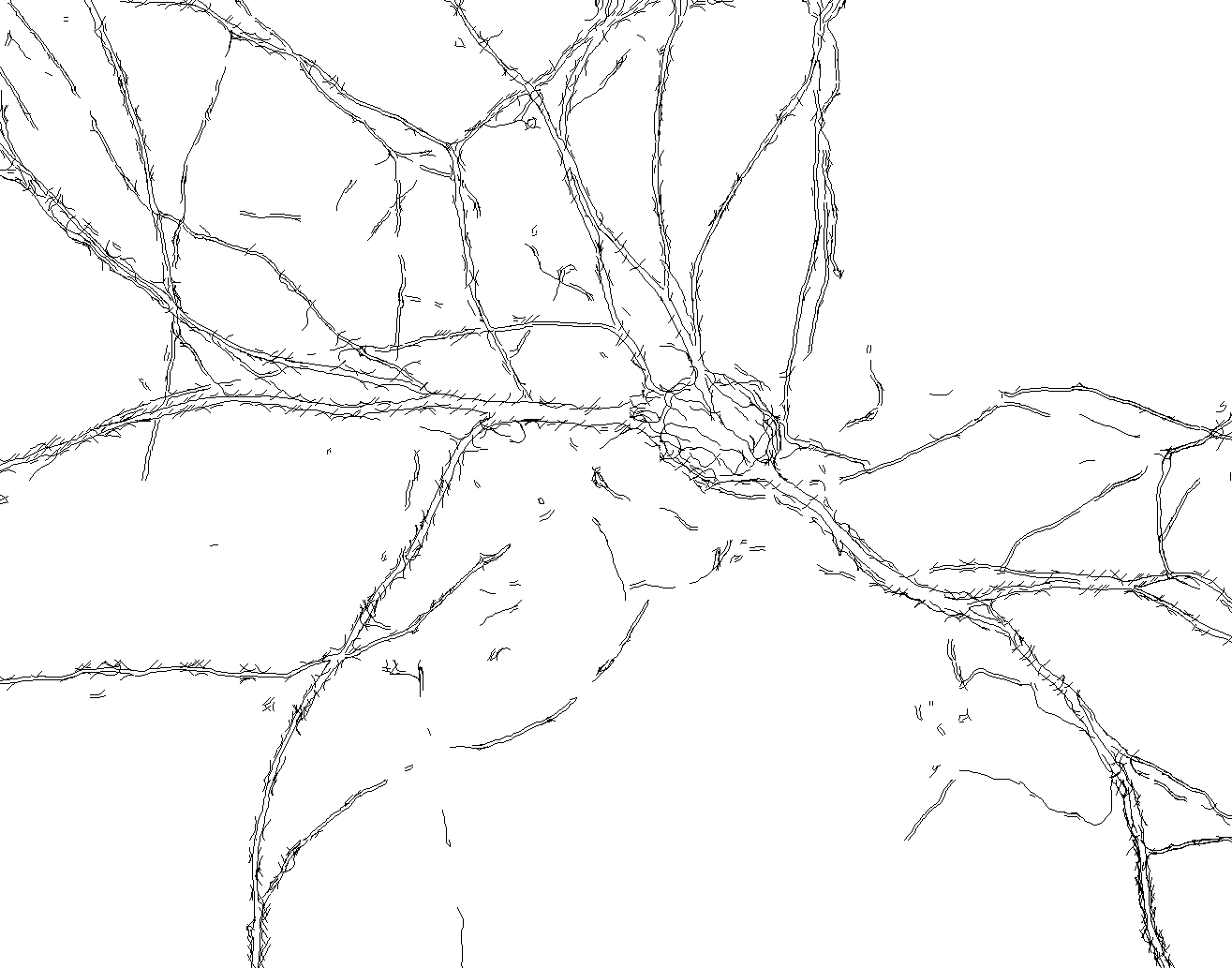}}
\caption{Example of a medical image with many curved edges, and the edge maps computed by deep        learning and classic approaches. (a) The original
	image. (b) The DL FED-CNN approach results \cite{ofir2018multi}. (c) The classic FastEdges
	\protect\cite{ofir2016fast} results. Both methods achieve high quality of
	detection while the DL runs in milliseconds and the classic runtime is more than
	seconds.}
\label{fig:example}
\end{figure*}

In order to conduct that investigation, we focus on two computer vision tasks that are at the limit of the ability of current state-of-the-art algorithms, i.e., faint edge detection in noisy images and multispectral registration of images.
Edge detection is one of the earliest problems that has been studied in image processing and computer vision \cite{sobel,marr1980theory,canny1987computational}. Although many approaches have been proposed to address this task, they still fail to detect edges when they are faint and the images are noisy \cite{ofir2016fast,ofir2019detection}. Those limitations are particularly problematic as these kinds of edges can be found in most imaging domains including satellite, medical, low-light, and even natural images. See Figure \ref{fig:example} for the classic and DL results of the faint edge detection methods that we discuss in this paper.

With the development of multi-sensor cameras that capture images from different modalities, multispectral image alignment has become a very important computer vision task. Indeed, robust alignment between the different image channels forms the basis for informative image fusion and data fusion. For example, while robust object detection can be derived from a combination of color and infrared images, this relies on the availability of accurate multispectral alignment. However, specialized methods need to be developed as reliable cross-spectral alignments cannot be achieved by using single-channel registration methods like scale-invariant feature transform (SIFT) \cite{lowe2004distinctive,brown2007automatic} feature based registration. 

Although a few comparative studies between DL and classic approaches have already been performed, this is the first that focuses on challenging ill-posed problems, exemplified by faint edge detection and multispectral image registration, which allow gaining interesting new insights. This paper is organized as follows. In Section \ref{sec:previous} we review previous studies analyzing classic and DL approaches. While in Section \ref{sec:fed} we compare such solutions for faint edge detection, in Section \ref{sec:registration} we focus on multispectral image alignment. Finally, we discuss the insights gained from this study in Section \ref{sec:Discussion} and conclude this manuscript in Section 
\ref{sec:conclusions}.

\section{Previous Work} \label{sec:previous}

Herein, 'classic' approaches are defined as those that do not depend on machine learning. They are engineered algorithms that rely on theory and mathematical models, and not directly on external data. 
Examples of such algorithms include: the Canny edge detector \cite{canny1987computational}, which uses hysteresis of gradients to identify curves in the image, the SIFT descriptor \cite{lowe2004distinctive}, which is an engineered and handcrafted representation of an image interest point, and optimization methods like in photometric stereo for example \cite{bartal2018photometric}.

A recent study \cite{classic-deep} compares classic and deep learning algorithms. That investigation focuses on three computer vision tasks, i.e., panorama generation, 3D reconstruction, and simultaneous localization and mapping (SLAM). They show that each approach, classic and DL, has its advantages and limitations. In particular, they highlight that the classic development process often relies on a strong theoretical framework which gives transparency and trust, whereas DL methods, when trained with an appropriate dataset, tend to deliver much higher performance. Other studies, focused on a single application, report outcomes of experiments evaluating their difference in terms of accuracy. A recent publication \cite{bojanic2019comparison} presents a comparison of a set of classic keypoint descriptors with their deep learning-based competitors \cite{lfnetono2018lf,superpoint-detone2018superpoint}. Evaluation under various geometric and illuminations shows that some combinations of classic keypoint detectors and descriptors outperform pre-trained deep models. On the other hand, performance analysis of two solutions for visual object detection, i.e., a classic feature extractor with a learned classifier and an object detector based on compact CNN (YOLO v3) \cite{redmon2018yolov3}, reaches a different conclusion \cite{manzoor2019comparison}. They find that the classic detector fails to detect objects under varying geometry such as size and rotations, while the compact CNN-based detector deals with these variations outperforming it. Similarly, a survey of classic and DL methods for face recognition \cite{trigueros2018face} confirms what is generally accepted in the community, like in boundary detection, e.g., \cite{hed}, that CNNs are the state of the art as they deliver significantly better accuracy. 

While performance metrics, such as accuracy, are key elements when comparing different approaches, researchers have also considered other aspects in their analysis. First, the high cost of the training phase of DL algorithms and its associated large amount of energy consumption have been highlighted \cite{hao2019training}. Second, evaluation of the computational resource requirements for DL, for NLP algorithms in particular, has drawn attention to the fact that, although large neural networks can improve accuracy, they rely on the availability of large and costly computational devices, which may limit their applicability \cite{strubell2019energy}. They report that training of an NLP standard DL model, like the one in \cite{bahdanau2014neural}, requires 120 training hours which can cost up to 180 USD of cloud computing and electricity. Third, a major limitation of current DL methods is the limited ability of humans to interpret them, i.e., the infamous black-box effect. This lack of transparency may prevent the deployment of DL-based solutions in applications where legal and ethical issues are paramount, such as autonomous driving \cite{lin2016ethics}.

Although previous research already provides good insight, further investigation is required in particular regarding assessing the behaviors of those classes of approaches when faced with challenging ill-posed problems. Thus, we conduct our research focusing on two tasks of that nature, i.e., faint edge detection and multispectral image registration, which are both long-standing research areas. We anticipate that the outcome of this study will inform the computer vision community about the ability of classic and DL methods to solve problems that are currently only addressed by weak solutions. Our comparison is discussed differently in detail in the thesis \cite{ofir2021classic}.

\section{Faint Edge Detection} \label{sec:fed}

\begin{figure}[tbh]
	\centering
	\includegraphics[width=0.99\linewidth]{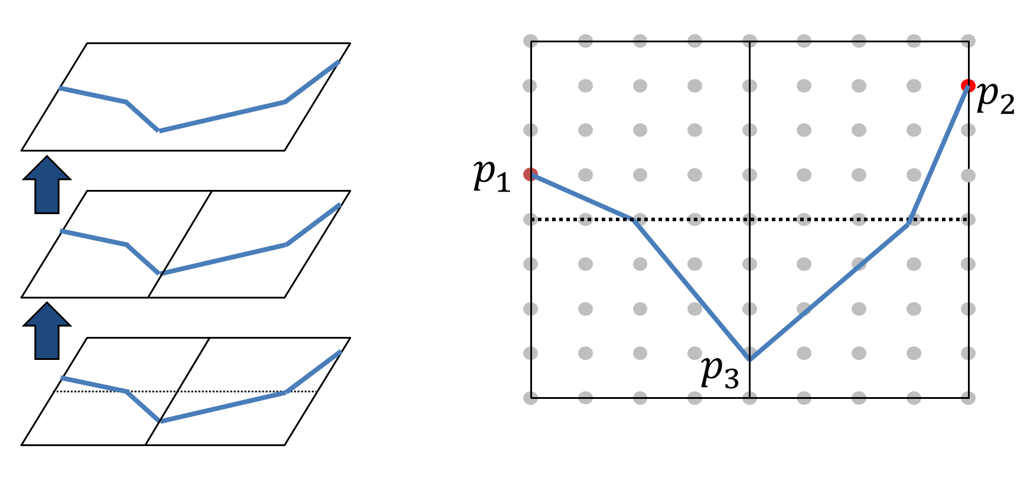}
	\caption{FED approach based on an image Rectangle-Partition-Tree  \cite{ofir2016fast}: this classic method searches the best concatenation of sub-curves by breaking point $p_3$ for every curve between every two boundary points $\forall p_1,p_2$. This search is performed recursively in a bottom-up dynamic programming-like approach.}
	\label{fig:RPT}
\end{figure}

\begin{figure}[tbh]
	\centering
	\includegraphics[width=0.99\linewidth]{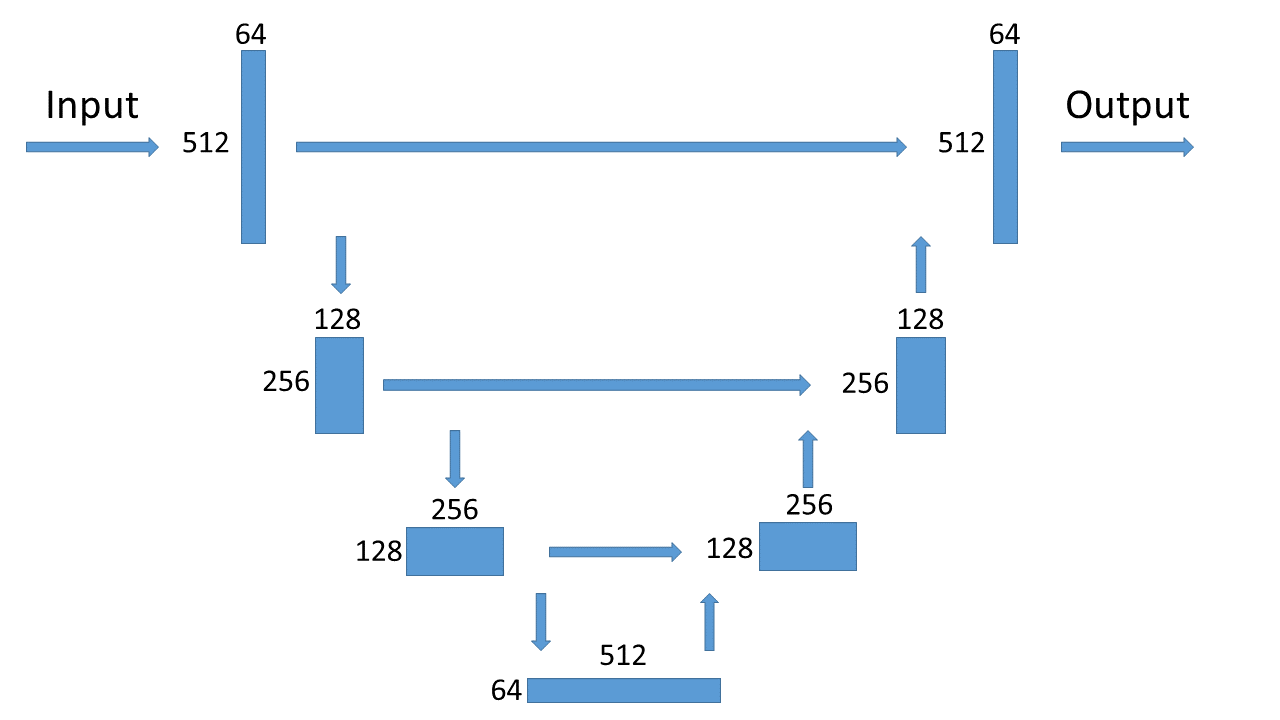}
	\caption{FED-CNN approach relying on a network of a U-Net Architecture \cite{ofir2018multi}: this multi scale CNN mimics the hierarchical tree approach of the classic algorithm \cite{ofir2016fast}. This deep neural network was trained using an edge preservation loss.}
	\label{fig:UNET}
\end{figure}

Faint edge detection (FED) is a challenging problem that has still not been addressed adequately. As edges at a low signal-to-noise ratio (SNR) can be found in a variety of domains, e.g., medical, satellite, low-light, and even natural images, effective solutions tend to be customized and applicable only to a very narrow range of applications \cite{crisp}. Recently, a couple of related state-of-the-art approaches have been proposed to improve FED accuracy: while FastEdges is a classic method relying on a hierarchical binary partitioning of the image pixels \cite{ofir2016fast} - see Figure \ref{fig:RPT}, FED-CNN takes advantage of a multiscale CNN to mimic that hierarchical tree approach \cite{ofir2018multi} - see Figure \ref{fig:UNET}. 

Using a simulation where a set of binary images \cite{binary} are contaminated by Gaussian additive noise and edges had their contrast reduced, we compare their performance highlighting their individual strengths and limitations. Note that the standard Canny detector \cite{canny1987computational} and the more recent Holistically Edge Detector (HED) \cite{hed}, a DL method based on the VGG-16 network \cite{Simonyan15}, are used as baseline methods. 
As it is common in the evaluation of binary classifiers, the F-measure, i.e., the harmonic mean of the precision and recall, is used to assess the quality of the detected edges. 

\begin{figure}[tbh]
	\centering
	\includegraphics[width=0.99\linewidth]{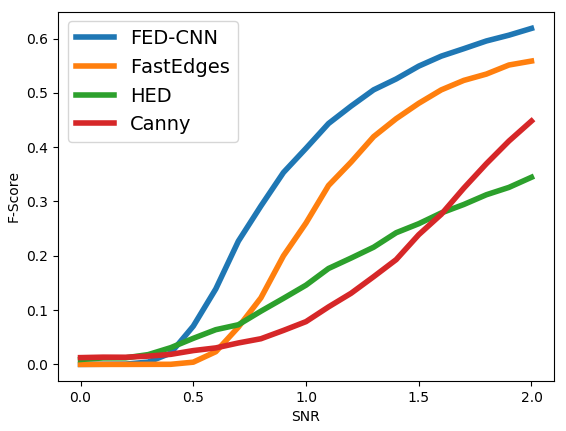}~
	\caption{Comparison of faint edge detectors: F-scores are calculated according to signal-to-noise ratios in the range [0,2].}
	\label{fig:snrGraph}
\end{figure}

As Figure \ref{fig:snrGraph} shows, where F-scores are calculated according to SNR, both state-of-the-art methods outperform significantly the HED and Canny detectors. However, FED-CNN systematically delivers higher F-scores than FastEdges's. For example, for a SNR of 1, resp. 2, FED-CNN achieves a score of 0.4, resp. 0.62, while FastEdges only obtains 0.28, resp. 0.56.

When considering computational complexity and runtime, again FED-CNN performs much better than FastEdges. First, a theoretical study of the computational complexity of those two algorithms reveals that, while FastEdges is nearly linear \cite{ofir2016fast}, FED-CNN is linear \cite{ofir2018multi}. Second, as Table \ref{table:runtime} shows, using an Intel i9 Sky-Lake CPU, FED-CNN proved more than 3 times faster than FastEdges. Moreover, the processing time of FED-CNN can easily be accelerated on a GPU, here a GeForce gtx1070, improving runtime by almost two orders of magnitude and approaching the speed of the efficient Canny detector.

\begin{table}[tbh]
	\centering%
	\caption{ Comparison of faint edge detectors according to runtime. Note that FED-CNN was implemented on both CPU and GPU hardware.}
	\begin{tabular}{|l|c|}
		\hline
		Algorithm & Runtime (in milliseconds) \\ \hline\hline
		FED-CNN (GPU) & \textbf{10} \\ \hline
		FED-CNN (CPU) & \textbf{800} \\ \hline
		FastEdges & 2600 \\ \hline
		Canny & 3 \\ \hline
	\end{tabular}%
	
	\label{table:runtime}
\end{table}

Although this experiment results in the DL algorithm outperforming significantly the classic one, the traditional approach has clear advantages. 

First, it has strong theoretic foundations \cite{ofir2016fast}. Its complexity, $C(N)$, can be expressed mathematically, where $N$ is the number of image pixels, and $l$ denotes the hierarchical level: 
\begin{equation}
C(N) \leq  6N^{1.5}\left[\sum_{l=0}^{\infty}2^{-l}+\sum_{l=1}^{\infty}2^{-l}\right]=18N^{1.5}
\end{equation}
Moreover, how faint an edge can be and still be detected by this classic algorithm is known. If $\sigma$ denotes the noise standard deviations and $w$ the filter width, the lower-bound of the minimal contrast that it can detect is: 
\begin{equation}
T_\infty = \Omega(\frac{\sigma}{\sqrt{w}})
\end{equation}
 This limit can be explained by i) the space of possible curves of the algorithm is exponential according to the curve length, and ii) the dynamic programming method used to search for an exponential number of curves takes a polynomial time.

Second, while differences between the natures of the training and testing sets generally lead to much-reduced performance of DL algorithms due to generalization bounds \cite{haussler1990probably}, classic methods tend to be suitable for various imaging domains. Indeed, although the design of FastEdges assumed step edges with constant contrast and Gaussian noise, this approach also achieved accurate results on the BSDS-500 \cite{martin2001database} dataset \cite{ofir2016fast}. This demonstrates that it can still be highly competitive in other imaging domains, such as those covered by BSDS-500 with its noisy natural images. On the other hand, when applied to an imaging domain similar to the training set's, FED-CNN shows high flexibility to geometric variations including edge curvatures and geometric transformations \cite{ofir2018multi}.

While performance scores are essential when selecting an approach, the cost of its development is also important. The development processes of classic and DL solutions are quite distinct. Whereas the FED classic approach required planning, analysis, parameter optimization, and complex derivation of computational complexity and threshold, the DL one, once suitable training data were identified, could be produced quite swiftly by adapting existing DL architectures. This versatility of DL architectures allows successful designs to be easily remodeled to address applications different from the ones for which they were initially conceived. As reported in \cite{ofir2018multi}, FED-CNN could be effortlessly transformed so that it could be used to perform noisy image classification and natural image denoising. Actually, experiments on the CIFAR 10 and 100 datasets \cite{cifar} revealed state-of-the-art accuracy \cite{ofir2018multi}.


\section{Multispectral Image Registration} \label{sec:registration}

\begin{figure*}
	\centering
	\includegraphics[width=140px]{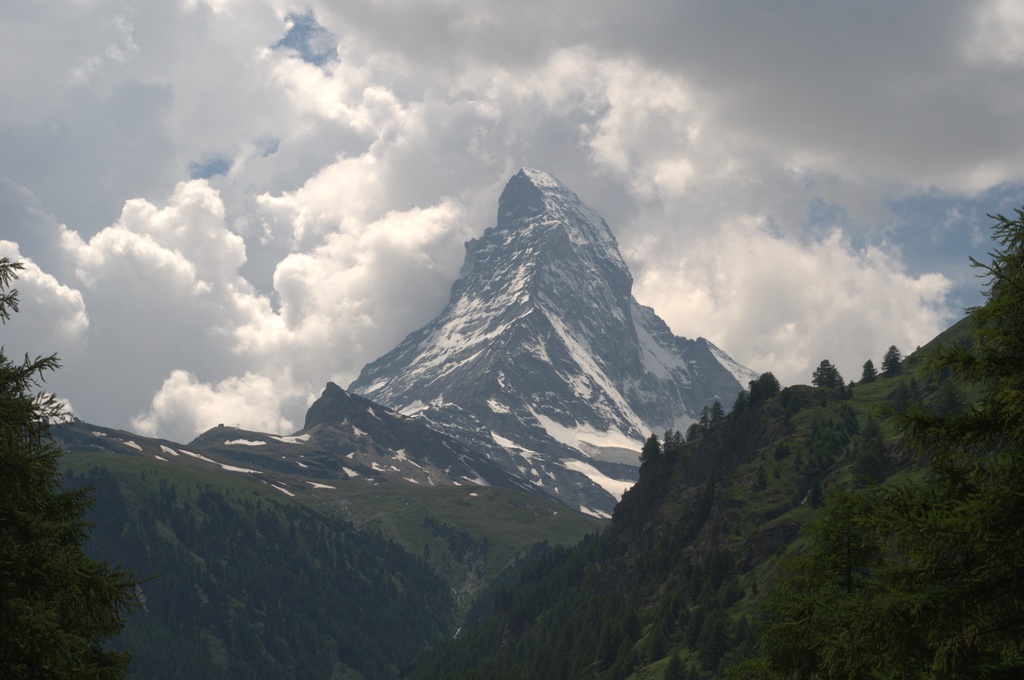}~
	\includegraphics[width=140px]{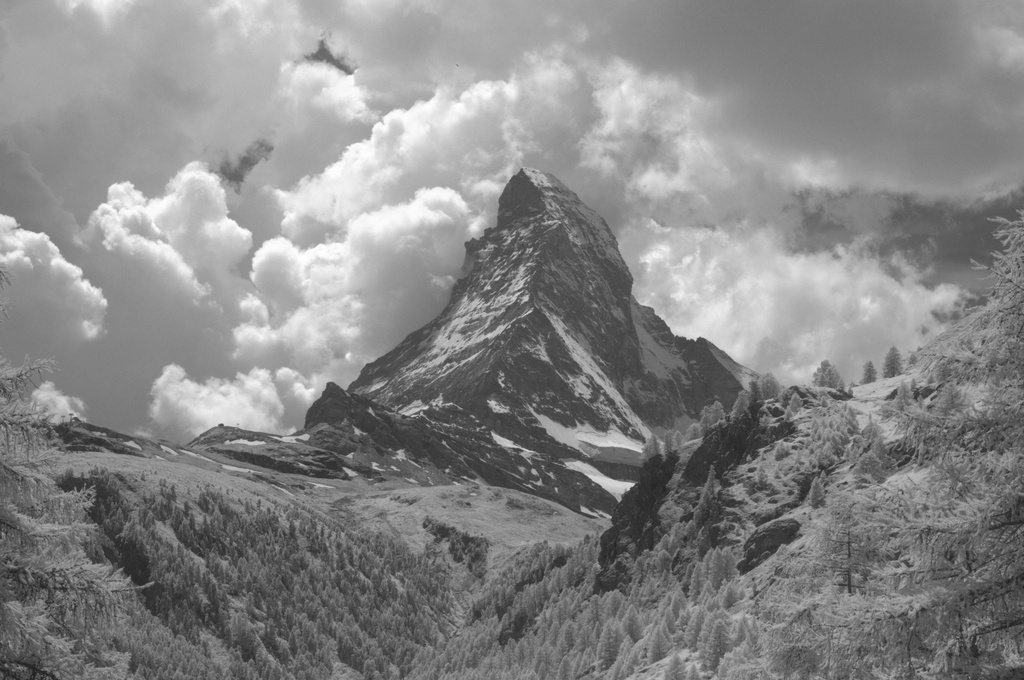}
	
	\caption{Pair of aligned cross-spectral images from the dataset used to train and evaluate the DL method \cite{ofir2018deep}. Left: RGB. Right: Near-Infra-Red (NIR).}
	\label{fig:rgbnir}
\end{figure*}

\begin{figure}[tbh]
	\centering
	\includegraphics[width=0.99\linewidth]{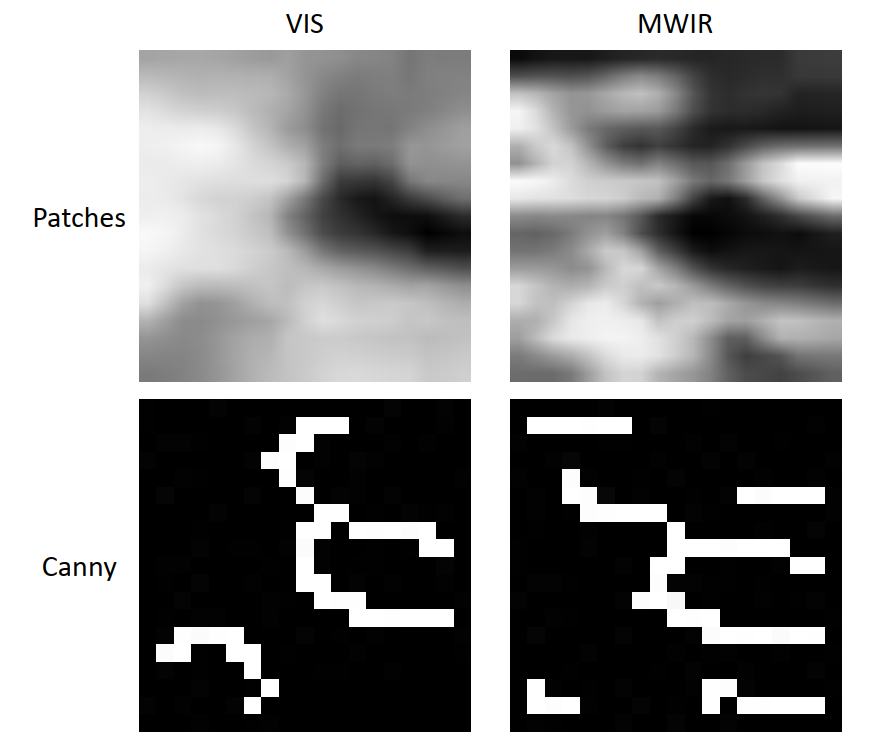}
	\caption{Multispectral patches and their corresponding edge maps that are part of their edge descriptors \cite{ofir2018registration}. Although the original patches are not correlated, their edge maps display significant similarity.}
	\label{fig:EdgeDescriptor}
\end{figure}

\begin{figure*}[tbh]
	\centering
	\includegraphics[width=1.0\linewidth]{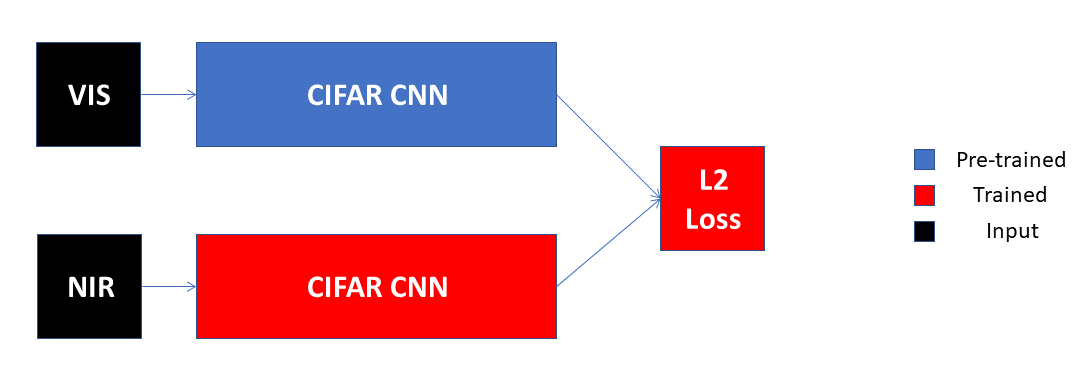}
	\caption{Deep-learning architecture for a learned invariant descriptor \cite{ofir2018deep}. It can be seen as a pseudo-Siamese network or as a teacher-student scheme. The visible (VIS) color patch is forwarded through a pre-trained classification network that was trained on the CIFAR10 dataset. Its corresponding Near-Infrared (NIR) patch is used to train the infrared network to produce a similar invariant representation. A L2 loss function is used.}
	\label{fig:learning}
\end{figure*}

Multispectral image alignment is another task which has not been satisfactorily addressed by computer vision. See Figure \ref{fig:rgbnir} for example of multispectral image pair. In this study, we focus on two recent developments which achieved consecutively state-of-the-art performance: a classic approach which relies on a handcrafted descriptor designed to be invariant to different spectra \cite{ofir2018registration} - see Figure \ref{fig:EdgeDescriptor} - and a DL framework based on pseudo-Siamese network \cite{ofir2018deep} - see Figure \ref{fig:learning}. 

Table \ref{table:registration} reports the average pixel error of those two approaches and other classic techniques, i.e., correlation of Canny \cite{canny1987computational}, correlation of Sobel \cite{sobel}, maximization of mutual-information and LGHD \cite{lghd2015}, in a task aiming at aligning visible (VIS), i.e., 0.4-0.7 $\mu m$, to Near-Infra-Red (NIR), i.e., 0.7-2.5 $\mu m$, images. This experiment was conducted using a standard dataset of cross-spectral aligned images \cite{BS11}.
The DL solution outperformed significantly all classic approaches. Moreover, as reported in \cite{ofir2018deep}, it is robust to geometric distortions: scaling applied in the [0.8,1.1] range only leads to a translation error of around 1 pixel.

\begin{table}[htb]
	\centering
	\caption{Error in pixels in a VIS to NIR image alignment task. The DL solution is the most accurate among all other methods.}
	\begin{tabular}{ l | c }
		Algorithm & Average pixel error \\
		\hline
		DL solution \cite{ofir2018deep} & \textbf{0.03} \\
		Handcrafted descriptor \cite{ofir2018registration} & 0.08 \\
		Canny & 0.07 \\
		Sobel & 0.07 \\
		Mutual Information & 0.11 \\
		LGHD & 0.21
	\end{tabular}
	\label{table:registration}
\end{table}

To evaluate if the DL approach was suitable to other imaging domains, it was applied on an alignment task of VIS to Middle-Wave Infrared (MWIR), i.e., $3-5\mu m$, images. However, since it had only been trained on a VIS to NIR dataset, this led to total failure. On the other hand, the application of the algorithm with its handcrafted descriptor to VIS to MWIR image alignment continued to deliver quality results \cite{ofir2018registration}, demonstrating the robustness of the classic approach to various spectral channels. 


As it has been seen, multispectral alignment can be performed using an approach either relying on a carefully crafted feature descriptor or learned by a CNN using a metric learning scheme. However, in terms of registration accuracy, while the DL approach excelled on images the features of which were covered in the training set and succeeded at handling geometric variations, the classic approach proved more robust to different imaging modalities. 

Although registration error is a key element when comparing multispectral image registration algorithms, other important aspects could also be considered. First, as the DL approach requires a forward pass of a CNN for every keypoint, the processing time of creating a feature descriptor is faster with the classic approach. Second, while a classic approach does not require training resources, the DL method relies on the availability of a valid multispectral database with a corresponding aligned image to operate. Moreover, its accuracy also depends on the level of information available in the keypoint features in that dataset. Third, both approaches have different hardware requirements: whereas the classic methods can easily be run on a standard CPU, real-time computing can only be achieved by the DL method if its execution takes place on a GPU. Not only is an expensive processing platform required, but also this prevents its usage on some embedded systems. Finally, there is a major difference regarding the development time that was needed to produce those two solutions. While the classic method was developed with much effort, once available, it could be quite rapidly transformed into its deep learning variant.

\section{Discussion} \label{sec:Discussion}

This comparison of recent classic and DL algorithms addressing two challenging ill-posed problems, i.e., faint edge detection and multispectral image registration, has provided novel insights regarding those two classes of approaches. Their particular features are summarized in Table \ref{table:comparison}. 

\begin{table}[tbh]
\centering
\caption{Comparison of the features of DL and classic approaches observed in this study.}
\begin{tabular}{ | l | l | l | }
\hline
 Feature/Approach & DL  & Classic\\ 
 \hline
 Accuracy (Acc.) & \textbf{High} & Moderate  \\  
 Acc. for other domains & Low & \textbf{Moderate} \\
 Speed on CPU & Slow & Slow/\textbf{Fast} \\
 Speed on GPU & Fast & / \\
 Theorical basis & Moderate & \textbf{High/Moderate} \\
 Training dataset & Essential & \textbf{No} \\
 Geometric variability & \textbf{Robust} & Weak \\
 Development & \textbf{Fast} & Slow \\
 Repurposing ability & \textbf{High} & Low \\
\hline
\end{tabular}
\label{table:comparison}
\end{table}

Although like most previous comparative studies \cite{manzoor2019comparison}, \cite{classic-deep}, and \cite{trigueros2018face}, ours reports that DL approaches achieve higher accuracy than classic methods. Moreover, it also underlines the fact that usage of a DL solution is very much restricted by the nature of its training set and, thus, it performs poorly when applied in another imaging domain. We should however note that they proved remarkably robust to geometrical transformations. As \cite{strubell2019energy}, our experiments also show that DL algorithms are slow on CPU-based machines, while they are appropriate for many classic solutions. Thus, GPU hardware is highly desirable when running DL solutions, which limits their applicability. Note that there are approaches of pruning and quantization that aim at minimizing inference time while preserving accuracy for DL \cite{han2015deep}.

Classic algorithms may be conceived from a strong theoretic basis, providing, e.g., in the case of faint edge detection, quantified information regarding the limit of their capacities. Unfortunately, as already mentioned by \cite{classic-deep}, this is not the case of the studied DL solutions, where, e.g., there is no practical understanding of either the CNN filter derived for FED or the invariant descriptor produced for multispectral image registration. This lack of transparency may prevent their usage in sensitive applications.

Since we had inside knowledge regarding the development of all the methods that we have investigated in this manuscript, we were in the quite unique position of being able to compare their development process. For both applications, once suitable training datasets were available, the implementation of the DL solution was much faster than the classic ones as existing CNN architectures could be quite easily adapted to fulfill the requirements of the targeted tasks. This repurposing ability can also naturally be exploited by recycling the DL algorithms investigated in this study. Indeed, FED-CNN was converted into both a noisy image classifier and a natural image denoiser by retraining the same CNN architecture using a different loss function.

Beyond accuracy, which, generally, privileges DL solutions if an appropriate training set can be assembled, we have reviewed other parameters that influence and sometimes impose the choice of a class of approaches when addressing computer vision and image processing applications. As both classes have unique strengths and limitations, it is expected that both will continue to produce useful solutions in the near future.

\section{Conclusion} \label{sec:conclusions}

In this paper, we reported the insights gained from a comparative study between DL and classic approaches applied to computer vision and image processing. In this investigation, we focused on challenging ill-posed problems, namely faint edge detection and multispectral image registration, analyzing the strengths and limitations of recent state-of-the-art DL and classic solutions.

Although those DL algorithms outperform classic methods in terms of accuracy and are robust to geometrical transformations, unlike the classic approaches, their performance collapses when attempting to process images outside their training space. Moreover, usage of GPUs is often mandatory to meet their generally higher computing requirements. On the other hand, the repurposing ability of DL architectures makes the development of new approaches much easier than with classic methods.

Eventually, the main concern regarding DL solutions may be that, while classic algorithms are quite transparent and are often supported by theory, the learning solutions are difficult to understand and explain. Thus,  until further progress in the interpretability of deep learning models, the issue of trust may hinder their deployment in many real-life applications. 

{\small
\bibliographystyle{ieee}
\bibliography{egbib}

\begin{thebibliography}{10}\itemsep=-1pt

\bibitem{lghd2015}
C.~Aguilera, A.~D. Sappa, and R.~Toledo.
\newblock Lghd: A feature descriptor for matching across non-linear intensity
  variations.
\newblock In {\em IEEE International Conference on Image Processing}, page~5.
  IEEE, Sep 2015.

\bibitem{bahdanau2014neural}
D.~Bahdanau, K.~Cho, and Y.~Bengio.
\newblock Neural machine translation by jointly learning to align and
  translate.
\newblock {\em arXiv preprint arXiv:1409.0473}, 2014.

\bibitem{bartal2018photometric}
O.~Bartal, N.~Ofir, Y.~Lipman, and R.~Basri.
\newblock Photometric stereo by hemispherical metric embedding.
\newblock {\em Journal of Mathematical Imaging and Vision}, 60(2):148--162,
  2018.

\bibitem{bojanic2019comparison}
D.~Bojani{\'c}, K.~Bartol, T.~Pribani{\'c}, T.~Petkovi{\'c}, Y.~D. Donoso, and
  J.~S. Mas.
\newblock On the comparison of classic and deep keypoint detector and
  descriptor methods.
\newblock In {\em International Symposium on Image and Signal Processing and
  Analysis}, pages 64--69. IEEE, 2019.

\bibitem{brown2007automatic}
M.~Brown and D.~G. Lowe.
\newblock Automatic panoramic image stitching using invariant features.
\newblock {\em International journal of computer vision}, 74(1):59--73, 2007.

\bibitem{BS11}
M.~Brown and S.~S\"usstrunk.
\newblock Multispectral {SIFT} for scene category recognition.
\newblock In {\em Computer Vision and Pattern Recognition (CVPR11)}, pages
  177--184, Colorado Springs, June 2011.

\bibitem{canny1987computational}
J.~Canny.
\newblock A computational approach to edge detection.
\newblock In {\em Readings in Computer Vision}, pages 184--203. Elsevier, 1987.

\bibitem{superpoint-detone2018superpoint}
D.~DeTone, T.~Malisiewicz, and A.~Rabinovich.
\newblock Superpoint: Self-supervised interest point detection and description.
\newblock In {\em Proceedings of the IEEE Conference on Computer Vision and
  Pattern Recognition Workshops}, pages 224--236, 2018.

\bibitem{sobel}
W.~Gao, X.~Zhang, L.~Yang, and H.~Liu.
\newblock An improved sobel edge detection.
\newblock In {\em IEEE International Conference on Computer Science and
  Information Technology}, volume~5, pages 67--71. IEEE, 2010.

\bibitem{han2015deep}
S.~Han, H.~Mao, and W.~J. Dally.
\newblock Deep compression: Compressing deep neural networks with pruning,
  trained quantization and huffman coding.
\newblock {\em arXiv preprint arXiv:1510.00149}, 2015.

\bibitem{hao2019training}
K.~Hao.
\newblock Training a single ai model can emit as much carbon as five cars in
  their lifetimes.
\newblock {\em MIT Technology Review}, 2019.

\bibitem{haussler1990probably}
D.~Haussler.
\newblock {\em Probably approximately correct learning}.
\newblock University of California, Santa Cruz, Computer Research Laboratory,
  1990.

\bibitem{crisp}
P.~Isola, D.~Zoran, D.~Krishnan, and E.~H. Adelson.
\newblock Crisp boundary detection using pointwise mutual information.
\newblock In {\em European Conference on Computer Vision}, pages 799--814.
  Springer, 2014.

\bibitem{cifar}
A.~Krizhevsky, V.~Nair, and G.~Hinton.
\newblock The cifar-10 dataset.
\newblock {\em online: http://www. cs. toronto. edu/kriz/cifar. html}, 2014.

\bibitem{binary}
L.~J. Latecki, R.~Lakamper, and T.~Eckhardt.
\newblock Shape descriptors for non-rigid shapes with a single closed contour.
\newblock In {\em Proceedings IEEE Conference on Computer Vision and Pattern
  Recognition}, volume~1, pages 424--429. IEEE, 2000.

\bibitem{lin2016ethics}
P.~Lin.
\newblock Why ethics matters for autonomous cars.
\newblock In {\em Autonomous driving}, pages 69--85. Springer, Berlin,
  Heidelberg, 2016.

\bibitem{lowe2004distinctive}
D.~G. Lowe.
\newblock Distinctive image features from scale-invariant keypoints.
\newblock {\em International journal of computer vision}, 60(2):91--110, 2004.

\bibitem{manzoor2019comparison}
S.~Manzoor, S.-H. Joo, and T.-Y. Kuc.
\newblock Comparison of object recognition approaches using traditional machine
  vision and modern deep learning techniques for mobile robot.
\newblock In {\em International Conference on Control, Automation and Systems},
  pages 1316--1321. IEEE, 2019.

\bibitem{marr1980theory}
D.~Marr and E.~Hildreth.
\newblock Theory of edge detection.
\newblock {\em Proc. R. Soc. Lond. B}, 207(1167):187--217, 1980.

\bibitem{martin2001database}
D.~Martin, C.~Fowlkes, D.~Tal, and J.~Malik.
\newblock A database of human segmented natural images and its application to
  evaluating segmentation algorithms and measuring ecological statistics.
\newblock In {\em Proceedings Eighth IEEE International Conference on Computer
  Vision}, volume~2, pages 416--423. IEEE, 2001.

\bibitem{ofir2019detection}
N.~Ofir, M.~Galun, S.~Alpert, A.~Brandt, B.~Nadler, and R.~Basri.
\newblock On detection of faint edges in noisy images.
\newblock {\em IEEE transactions on pattern analysis and machine intelligence},
  42(4):894--908, 2019.

\bibitem{ofir2016fast}
N.~Ofir, M.~Galun, B.~Nadler, and R.~Basri.
\newblock Fast detection of curved edges at low snr.
\newblock In {\em Proceedings of the IEEE Conference on Computer Vision and
  Pattern Recognition}, pages 213--221, 2016.

\bibitem{ofir2018multi}
N.~Ofir and Y.~Keller.
\newblock Multi-scale processing of noisy images using edge preservation
  losses.
\newblock In {\em International Conference on Pattern Recognition}, pages 1--8.
  IEEE, 2021.

\bibitem{ofir2018deep}
N.~Ofir, S.~Silberstein, H.~Levi, D.~Rozenbaum, Y.~Keller, and S.~D. Bar.
\newblock Deep multi-spectral registration using invariant descriptor learning.
\newblock In {\em IEEE International Conference on Image Processing}, pages
  1238--1242. IEEE, 2018.

\bibitem{ofir2018registration}
N.~Ofir, S.~Silberstein, D.~Rozenbaum, Y.~Keller, and S.~D. Bar.
\newblock Registration and fusion of multi-spectral images using a novel edge
  descriptor.
\newblock In {\em IEEE International Conference on Image Processing}, pages
  1857--1861. IEEE, 2018.

\bibitem{ofir2021classic}
Y.~N. Ofir.
\newblock {\em Classic versus deep learning approaches to address computer
  vision challenges: a study of faint edge detection and multispectral image
  registration}.
\newblock PhD thesis, Kingston University, 2021.

\bibitem{classic-deep}
N.~O’Mahony, S.~Campbell, A.~Carvalho, S.~Harapanahalli, G.~V. Hernandez,
  L.~Krpalkova, D.~Riordan, and J.~Walsh.
\newblock Deep learning vs. traditional computer vision.
\newblock In {\em Science and Information Conference}, pages 128--144.
  Springer, 2019.

\bibitem{lfnetono2018lf}
Y.~Ono, E.~Trulls, P.~Fua, and K.~M. Yi.
\newblock Lf-net: learning local features from images.
\newblock In {\em Advances in neural information processing systems}, pages
  6234--6244, 2018.

\bibitem{redmon2018yolov3}
J.~Redmon and A.~Farhadi.
\newblock Yolov3: An incremental improvement.
\newblock {\em arXiv preprint arXiv:1804.02767}, 2018.

\bibitem{Simonyan15}
K.~Simonyan and A.~Zisserman.
\newblock Very deep convolutional networks for large-scale image recognition.
\newblock In {\em International Conference on Learning Representations}, 2015.

\bibitem{strubell2019energy}
E.~Strubell, A.~Ganesh, and A.~McCallum.
\newblock Energy and policy considerations for deep learning in nlp.
\newblock {\em arXiv preprint arXiv:1906.02243}, 2019.

\bibitem{trigueros2018face}
D.~S. Trigueros, L.~Meng, and M.~Hartnett.
\newblock Face recognition: From traditional to deep learning methods.
\newblock {\em arXiv preprint arXiv:1811.00116}, 2018.

\bibitem{hed}
S.~Xie and Z.~Tu.
\newblock Holistically-nested edge detection.
\newblock In {\em Proceedings of the IEEE international conference on computer
  vision}, pages 1395--1403, 2015.

\end{thebibliography}
}

\end{document}